\pdfoutput=1

\documentclass[11pt]{article}

\usepackage{amsmath,amsfonts,bm}

\def\eqref#1{equation~\ref{#1}}

\def\1{\bm{1}}

\DeclareMathAlphabet{\mathsfit}{\encodingdefault}{\sfdefault}{m}{sl}
\SetMathAlphabet{\mathsfit}{bold}{\encodingdefault}{\sfdefault}{bx}{n}

\usepackage{acl}
\usepackage{times}
\usepackage{latexsym}
\usepackage{graphicx}
\usepackage[normalem]{ulem}
\usepackage[T1]{fontenc}

\usepackage[utf8]{inputenc}

\usepackage{microtype}
\usepackage{graphicx}
\usepackage{xspace}
\usepackage{tabularx}
\usepackage{booktabs}
\usepackage{multirow}
\usepackage{comment}
\usepackage{caption}
\usepackage{subcaption}
\usepackage{textgreek}
\usepackage{amssymb}
\usepackage{amsmath}
\usepackage{newunicodechar}
\usepackage{color}
\usepackage{tipa}
\usepackage[T4,T1]{fontenc}
\usepackage{color, colortbl}
\usepackage{siunitx}
\usepackage{pifont} 

\title{Geographical Distance Is The New Hyperparameter:\\ { A Case Study Of Finding The Optimal Pre-trained Language For English-isiZulu Machine Translation.}}

\author{\normalsize Muhammad Umair Nasir$^{1}$, Innocent Amos Mchechesi$^{2}$ \\\\
  \footnotesize
  $^{1}$ Ominor AI, $^{2}$ University of the Witwatersrand, South Africa}

\begin{document}
\maketitle
\begin{abstract}
Stemming from the limited availability of datasets and textual resources for low-resource languages such as isiZulu, there is a significant need to be able to harness knowledge from pre-trained models to improve low resource machine translation. Moreover, a lack of techniques to handle the complexities of morphologically rich languages has compounded the unequal development of translation models, with many widely spoken African languages being left behind. This study explores the potential benefits of transfer learning in an English-isiZulu translation framework. The results indicate the value of transfer learning from closely related languages to enhance the performance of low-resource translation models, thus providing a key strategy for low-resource translation going forward. We gathered results from 8 different language corpora, including one multi-lingual corpus, and saw that isiXhosa-isiZulu outperformed all languages, with a BLEU score of 8.56 on the test set which was better from the multi-lingual corpora pre-trained model by 2.73. We also derived a new coefficient, \emph{Nasir's Geographical Distance Coefficient (NGDC)} which provides an easy selection of languages for the pre-trained models. NGDC also indicated that isiXhosa should be selected as the language for the pre-trained model.
\end{abstract}

\section{Introduction}

Neural machine translation aims to automate the translation of text or speech from one language to another utilising neural networks \citep{nyoni2021low}. Consequently, the performance of neural machine translation (NMT) models is highly dependent on the availability of large parallel corpora to provide sufficient training data. Low-resource languages which are under-represented in internet sources lack suitable training corpora and therefore suffer from limited development, obtaining poor translation performance. This phenomenon is exacerbated by a lack of content creators, dataset curators and language specialists, resulting in barriers at many stages in the translation process \citep{lakew2020low,zoph2016transfer,sennrich2019revisiting}.

Therefore, due to the historical focus on dominant languages such as English in the development of neural machine translation (NMT) models, low-resource and morphologically complex languages remain a challenge for current translation systems \citep{haddow2021survey,koehn2017six}. Due to limited resources in terms of both computational expense and available datasets, it is vital to be able to leverage knowledge from current pretrained models to provide more effective solutions. Therefore, in this investigation, the effects of transfer learning from closely related languages, as well as comparison with high-resourced languages for pre-trained scenario, is explored in the context of English to Zulu translation.

Furthermore, this study derives the Nasir's Geographical Distance coefficient. \emph{Geographical Distance (GD)} \citep{holman2007relation} has been studied for various scientific research areas \citep{bei2021motivations,krajsa2011rtt,riginos2001population} as it provides deep insights in many aspects. We will also use GD as a hyperparameter for an attempt to get a language for a pre-trained model in an effective and with a $O(n)$ complexity. Although there are many ways to find GD, we will use literal approximation of distance in kilometers and suggest the techniques in future directions.




\subsection{Background}

Previous studies have indicated poor translation performance for the isiZulu languages due to its morphological complexity and limited available data \citep{martinus2019focus}. The challenging nature of English-isiZulu translation is highlighted in a benchmark of five low-resource African languages by \citet{martinus2019focus}, where isiZulu obtains a much poorer BLEU score in comparison to other evaluated languages. The study suggests that the collection of higher quality datasets for isiZulu would greatly benefit translation performance. 

Furthermore, the challenges associated with the morphological complexity of Nguni languages such as isiZulu are tackled in a study by \citet{moeng2021canonical}. The investigation explores the use of supervised sequence-to-sequence models to tokenize isiZulu, isiXhosa, isiNdebele and siSwati sentences, demonstrating promising results for improved segmentation of morphologically complex Nguni languages.

A notable study by \citet{nyoni2021low} compares the use of zero-shot learning, transfer learning and multi-lingual learning on three Bantu languages, namely isiZulu, isiXhosa and chiShona. The results indicate that multi-lingual learning where a many-to-many model was trained using three different language pairs, English-isiZulu, English-isiXhosa and isiXhosa-isiZulu led to optimal results on their custom dataset.

In addition, the study found that transfer learning from a closely related Bantu language is highly effective for low resource translation models, with statistically significant results being obtained when transfer learning to isiZulu using the pretrained English-to-isiXhosa model \citep{nyoni2021low}. In contrast, transfer learning from the English-to-Shona model did not yield any statistically significant improvement, indicating the role of morphological similarity in the transfer learning process. 

There has been a lot of work in providing assistance to low-resourced languages for machine translation focus of the area. \citet{neubig2018rapid} trained multilingual models as seed models and then continued training on low-resourced language. \citet{sennrich2015improving} looks into training monolingual data with automatic back-translation \citep{edunov2018understanding,caswell2019tagged,edunov2019evaluation} to improve scores through only a mono-lingual data. Another work that utilizes back-translation for effecctive NMT training is done by \citet{dou2020dynamic}. \citet{koneru2022cost} proposes a cost-effective training procedure to increase the performance of models on NMT tasks, utilizing a small number of annotated sentences and dictionary entries. \citet{park2020decoding} looked into decoding strategies for low-resourced languages in an attempt to improve training. \citet{nguyen2017transfer} looked into related languages to a target language for low-resourced languages to prove effectiveness of similar languages.

Similarly, this study aims to investigate whether transfer learning from a morphologically similar language will be effective on the novel, high-quality Umsuka English-isiZulu parallel corpus and if so, how does it perform when we use high-resourced mono- and multi-lingual corpora. This study will also derive a formula which will ease the way for selecting a language for a pre-trained model.

\section{Methodology}

This investigation evaluates several models pre-trained on different language pairs, both low- and high-resourced, on a recently release English-Zulu parallel corpus. The dataset utilized to fine-tune and benchmark the models is discussed below. 

\subsection{Dataset}

The Umsuka English-isiZulu Parallel Corpus \citep{zulu_corpus} provides a novel, high-quality parallel dataset for machine translation, containing English sentences sampled from both News Crawl datasets which were then translated into isiZulu, and isiZulu sentences from the NCHLT monolingual corpus and UKZN isiZulu National monolingual corpus, which were then translated into English. Each translation was performed twice, by two differing translators, due to the high morphological complexity of the isiZulu language. This also serves the purpose of considering one translation as a reference and the other as target. This can be validated as both have been translated by human annotators and are different from each other. The dataset is publicly available from the Zenodo platform\footnote{\url{https://zenodo.org/record/5035171##.YZvn1fFBy3J}}.

\subsection{Models}

The three models tested are based on the MarianMT model \citep{junczys2018marian} which is constructed using a Transformer architecture. Each model is pretrained on a different set of language pairs from the Helsinki Corpus.

\emph{MarianMT} \citep{junczys2018marian} is a toolkit for neural machine translation written in C++ with over 1000 models trained on different language pairs from OPUS\footnote{\url{https://opus.nlpl.eu/}}, available at the HuggingFace Model Hub\footnote{\url{https://huggingface.co/}}. Each model is based on a Transformer encoder-decoder structure with 6 layers in each component \citep{junczys2018marian}. From the available models, $8$ pre-trained models were selected\footnote{\url{https://github.com/umair-nasir14/NGDC}}, representing pre-training on a closely related language, pre-training on a more distantly related language within the same family and pre-training on multiple unrelated languages, with less and more data, respectively. Since each model was based on the same architecture, this allowed for a controlled comparison of the language pairs used for pre-training,  as any discrepancies due to architectural differences were discounted.

Since  isiXhosa and isiZulu are both part of the Nguni branch of Bantu languages, isiXhosa is closely related to isiZulu in the Bantu language family tree \cite{nyoni2021low}. As well as Shona, or chiShona, is selected as it is also a part of Southern Bantu language group \citep{nyoni2021low}. Another Bantu language, Kiswahili was explored to determine the effects of transfer learning from another language within the Bantu family which is not as closely related to the target isiZulu language. While isiZulu is classified as a Southern Bantu and Nguni language, Kiswahili is part of the Northeast Bantu and Sabaki languages \cite{nurse1993swahili}.

Twi, or Akan-kasa, is spoken in Ghana, has been selected to have a representation from Western Africa and to explore the effects a dialect of the Akan language on fine-tuning isiZulu. Luganda is selected as a representation from Niger-Congo family of languages and is spoken in East-African Country of Uganda. This will able us to explore the fine-tuning regime in Niger-Congo languages.

Arabic and French are selected as they are morphologically very different and are considered to be high-resourced \citep{ali2014advances,besacier2014introduction}. We explore effects of fine-tuning high-resourced languages with different morphologies. As the notion of having more and multi-lingual data will be better for fine-tuning, we select a corpus of Romance languages, which is created by joining 48 Romance languages including French, Italian, Spanish, Walloon, Catalan, Occitan, Romansh etc. We include Romance languages so that we can cover the aspect of big multi-lingual corpora being fine-tuned on low-resourced isiZulu and to prove our hypothesis.


\subsection{Implementation Reproducibility}

We believe all experiments must be \emph{Reproducible}. To achieve this we are open-sourcing our code on GitHub (added in the footnote previously).

\section{Results}

Each model was benchmarked on the test set using the BLEU\citep{papineni2002bleu} score as tabulated in Table \ref{table1} below. It can be observed that the optimal model is given by the MarianMT model pre-trained on the English-Xhosa dataset. This confirms our hypothesis that transfer learning from a geographically distant language would result in poor performance. Here GD is in Kilometers (Km) and corpus size is in Number Of Sentences in millions (M).


\begin{table*}
\centering

\begin{tabular}{|l|l|l|l|l|}
\hline
\textbf{Language(s)}        & \textbf{BLEU(Val)} & \textbf{BLEU(Test)} & \textbf{Corpus Size(NOS)} & \textbf{GD(KM)} \\ \hline
\textit{\textbf{isiXhosa}}  & 10.20                & 8.56                & 20.7                    & 1000              \\ \hline
\textit{\textbf{Romance}}   & 7.76                 & 5.83                & 1232.7                  & 13094.4           \\ \hline
\textit{\textbf{Arabic}}    & 5.76                 & 3.07                & 102.8                   & 5205              \\ \hline
\textit{\textbf{French}}    & 5.42                 & 3.91                & 479.1                   & 13094             \\ \hline
\textit{\textbf{Kiswahili}} & 5.28                 & 3.97                & 9.1                     & 3783.1            \\ \hline
\textit{\textbf{chiShona}}  & 4.32                 & 2.83                & 0.1                     & 1584              \\ \hline
\textit{\textbf{Twi}}       & 1.91                 & 1.34                & 0.047                   & 7962            \\ \hline
\textit{\textbf{Luganda}}   & 0.94                 & 0.55                & 0.039                   & 4883.7            \\ \hline
\end{tabular}
\caption{BLEU scores, GD and corpora size}
\label{table1}
\end{table*}

\begin{figure*}[h!]
\centering
\includegraphics[width=1.0\linewidth]{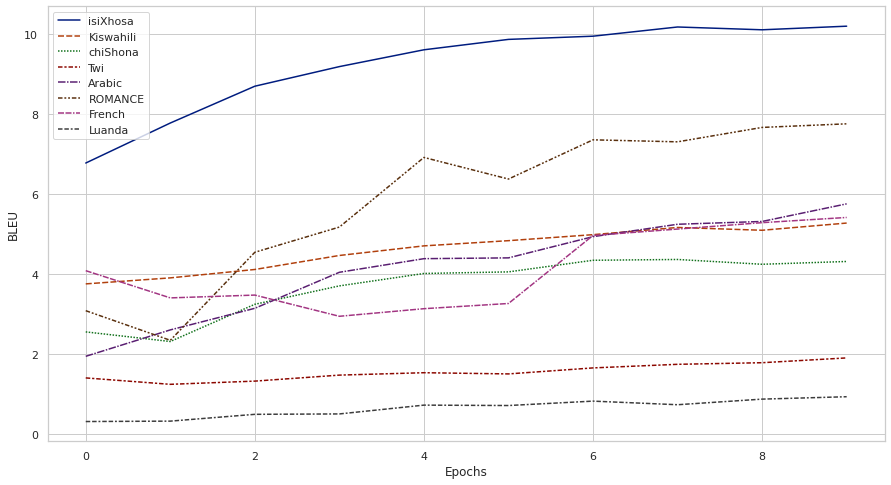}
\caption{BLEU scores per epoch according to different pre-training languages, indicates high performance of morphologically similar isiXhosa, which outperforms a model trained on a very large corpora and rest of corpora.}
\label{resultfig1}
\end{figure*}

In Fig. \ref{resultfig1} below, we can observe that the MarianMT model pre-trained on the English-Xhosa dataset outperforms all other models by a good margin, obtaining a final BLEU score of $8.56$. This result suggests that the morphological similarities between the isiZulu and isiXhosa languages plays a strong role in the benefits attained through fine-tuning.

Following identification of the optimal model, the MarianMT model pre-trained on the En-Xh dataset was further fine-tuned for 75 epochs on Umsuka dataset, giving a final optimal BLEU score of 17.61 on training set and 13.73 on test.

\section{Analysis}

We now present an analysis of the results in light of both the underlying theory and previous literature. In order to further understand the effects of pre-training on different languages, the datasets used for pre-training of the MarianMT models were inspected. Notably, although the number of sentences in English-Xhosa dataset is in order of magnitudes less than Romance languages corpus but still performs better. This justifies our hypothesis and opens up a path to effective fine-tuning through the knowladge of morphologies and not by adding multiple languages into a single corpus. Arabic and French having approximately $5$ and $23$ times more data also suggests the above mentioned hypothesis that with closer GD and lesser data is much better, in many ways, than larger data and farther GD.

Other Bantu languages that were selected, Kiswahili and chiShona performed almost similar to Arabic and French with order of magnitudes of lesser data which suggests that even if they are not as similar to isiZulu, the distance being very close to where isiZulu is spoken tends to have a great impact. We speak similar languages in neighbouring cities and countries which should have an effect on the model and so the result suggests. Twi and Luganda, having very less data and higher GD, gives us very poor results.

From Table \ref{table1}, we also observe that distance between the target language and the language from a pre-trained model is a very important factor. Alone, to a good extent it can serve the purpose of choosing the language of pre-trained model but we want to look one step deeper as one can argue that Romance languages corpora, French and Arabic perform relatively better but the distances are larger. Thus we also look into Size of Corpus (Table \ref{table1}). Which forces us to think about deriving a relationship that involves both distance and the size. This will be explained in the upcoming sub-section.

\begin{figure*}[h!]
\centering
\includegraphics[width=1.0\linewidth]{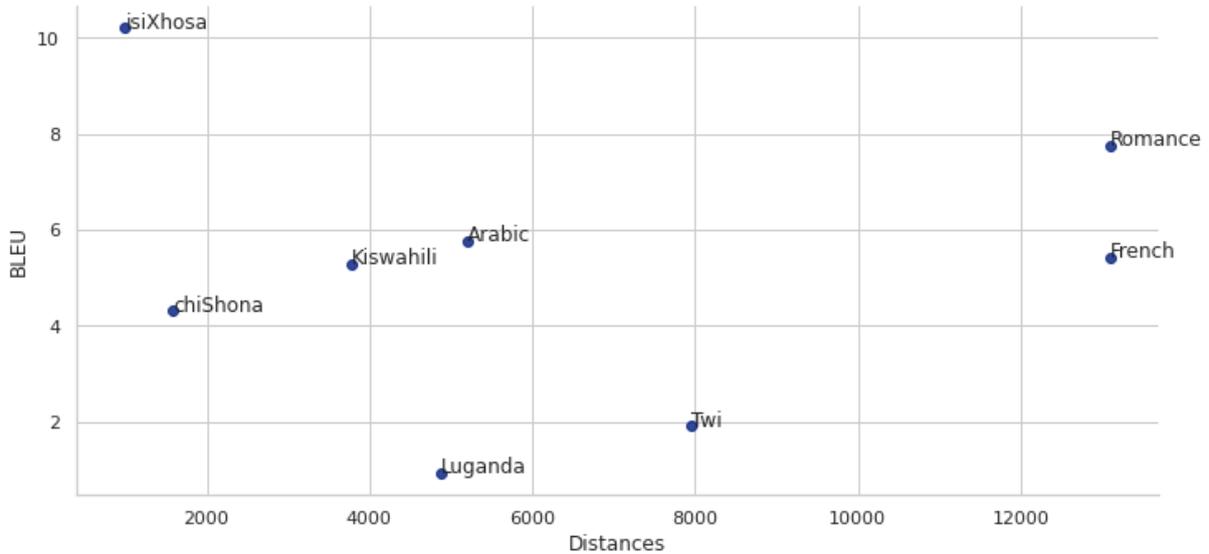}
\caption{Relationship between BLEU scores and distance (KM) of places where languages are spoken from the place where isiZulu is spoken. }
\label{resultfig2}
\end{figure*}

\subsection{Nasir's Geographical Distance Coefficient}

In Figure \ref{resultfig2} we can observe that there is a sensible relationship between BLEU scores and distance, and as a rule of thumb there should always be a relationship with corpus size \citep{lin2019choosing}. With further analysis we can deduce that neither distance alone nor corpus size alone can be taken for granted when selecting a language for pre-trained model. Thus, we derive a formula which takes into account both distance and corpus size in account. This formula is intended to be used before training to know which language corpora to select.
\\
\begin{equation*}
    z = \frac{cD}{(1-c)S}
\end{equation*}

\[
    \delta = 
\begin{cases}
    1 ,& \text{if } D\geq D_{max}\\
    \frac{exp^(z)}{1 + exp^(z)} ,  & \text{otherwise}
\end{cases}
\]

where $D$ is the distance between language to fine-tune and language of the pre-trained model, $S$ is the size of corpus, $c$ is the weight coefficient, set to $0.4$, which could act as hyperparameter. $D_{max}$ is also hyperparameter to be tuned when it is being used in different languages in different parts of the world. $\delta$ is the coefficient we are introducing, \emph{Nasir's Geographical Distance Coefficient (NGDC)}. The goal here is to minimize NGDC. \\

\begin{figure*}[h!]
\centering
\includegraphics[width=1.0\linewidth]{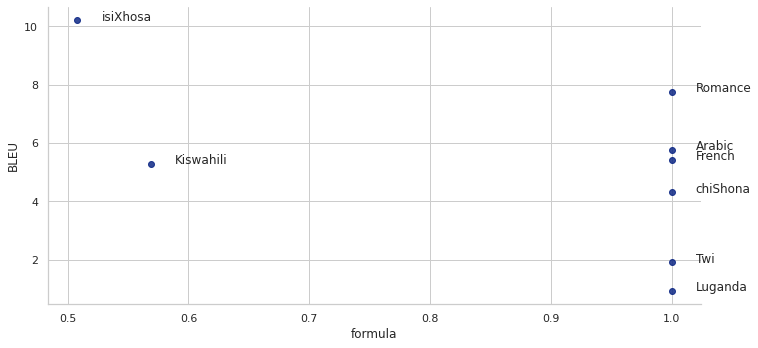}
\caption{NGDC with Penalty}
\label{withPen}
\end{figure*}

\begin{figure*}[h!]
\centering
\includegraphics[width=1.0\linewidth]{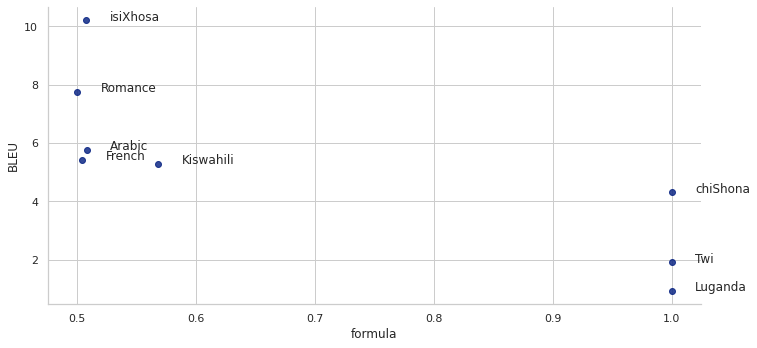}
\caption{NGDC without Penalty}
\label{withoutPen}
\end{figure*}

Table \ref{table:table2}, Figures \ref{withPen} and \ref{withoutPen} shows the results and effectiveness of our introduced NPC. We can observe that without imposing penalty we have Romance languages, Arabic and French as desired pre-trained model languages along with isiXhosa and Kiswahili, which makes absolute sense as some have more data and others are near to target language but we want to have morphologically closer languages which will get better results. It would also be better if lesser carbon footprint is left and lesser training resources are used. Thus, with the penalty we only get isiXhosa and Kiswahili as desired ones, which will eventually be better in all perspectives.

\begin{table*}
\centering

\begin{tabular}{|l|l|l|l|}
\hline
\textbf{Language(s)}        & \textbf{BLEU} & \textbf{NGDC(With Penalty)} & \textbf{NGDC(Without Penalty)} \\ \hline
\textit{\textbf{isiXhosa}}  & 10.20         & 0.5080                     & 0.5080                        \\ \hline
\textit{\textbf{Romance}}   & 7.76          & 1.0000                     & 0.5007                        \\ \hline
\textit{\textbf{Arabic}}    & 5.76          & 1.0000                     & 0.5084                        \\ \hline
\textit{\textbf{French}}    & 5.42          & 1.0000                     & 0.5045                        \\ \hline
\textit{\textbf{Kiswahili}} & 5.28          & 0.5688                     & 0.5688                        \\ \hline
\textit{\textbf{chiShona}}  & 4.32          & 0.9999                     & 0.9999                        \\ \hline
\textit{\textbf{Twi}}       & 1.91          & 1.0000                     & 1.0000                        \\ \hline
\textit{\textbf{Luganda}}   & 0.94          & 1.0000                     & 1.0000                        \\ \hline
\end{tabular}
\caption{NGDC with and without Penalty.}
\label{table:table2}
\end{table*}

\section{Impact Statement}

The potential impacts of this investigation can be explored in light of the possible contributions, risks and societal impact.

\subsection{Applications and Benefits}

The study poses potential benefits to further research into low-resource languages as it motivates careful choice of the pre-trained model used for transfer learning in order to improve performance on low resource languages. This could provide a vital tool to improve the efficiency and performance of low resource translation pipelines, especially in resource-constrained environments. In addition, this principle could be applied more broadly to other language groups with morphologically similar languages.

Moreover, effective transfer learning provides the additional advantage of promoting decreased computational expense since prior knowledge from previously trained networks can be leveraged effectively. This could work to mitigate the substantial detrimental environmental impact stemming from the intensive GPU training required to train neural machine translation models. This is critical to ensure sustainable development of machine translation models by minimising resource waste.

\subsection{Limitations and Drawbacks}

It should be noted that any conclusions drawn from the study are based on the BLEU score as the sole evaluation metric. This may provide a limited view of the true translation performance as it is based on n-gram similarity and does not necessarily measure whether the meaning of a sentence has been captured. A further improvement could be to conduct a similar study with additional expertise from a linguistic specialist to verify whether the output of the translation models is valid.

\subsection{Social Impact}

Societal impacts of low resource neural machine translation include furthering accessibility of information to under-represented languages and working to close the digital divide between high-resource and low-resource languages. Machine translation is an essential component of applications ranging from voice-assisted smart-phone applications that provide healthcare to rural communities to ensuring multi-lingual access to educational materials. Therefore it is vital that machine translation technology is accessible and functional for low-resource languages to be able to build valuable tools which could have a beneficial societal impact.

\section{Conclusion and Future Directions}

English-isiZulu translation has historically obtained poor results on translation benchmarks due to a lack of high-quality training data and appropriate tokenization schemes able to handle the agglutinative structure of isiZulu sentences. In this investigation, the challenges of isiZulu translation in terms of both morphological complexity and a lack of textual resources are explored using the recently released Umsuka English-isiZulu Parallel Corpus. In order to investigate the effects of  the impact of the pre-trained model selected for transfer learning, several models were fine-tuned and benchmarked on the Umsuka dataset.

MariantMT models pre-trained on English-Xhosa, English-Swahili, English-Shona, English-Twi, English-Luganda, English-Arabic, English-French and English-Multilingual Romance languages, respectively. The study found that the pre-trained English-Xhosa model attained the optimal results with a handsome margin. Thus, the results indicate that transfer learning is particularly effective when languages are within the same sub-family while transfer learning is less effective when the model is pre-trained on a more distantly related language, no matter the size of the data to an extreme extent. We have also introduced a novel \emph{Nasir's Geographical Distance Coefficient} which will help researchers find a language for pre-trained model effectively and will result in using less resources.

Therefore, this study motivates careful choice of the pre-trained model used for transfer learning, utilising existing knowledge of language family trees, to promote improved performance of low resource translation. In addition, we have open-sourced\footnote{\url{https://huggingface.co/MUNasir/umsuka-en-zu}} our best model which was fine-tuned for 75 epochs using the original MarianMT model pre-trained on the English-Xhosa language pair, obtaining a final BLEU score of 17.61 on train while 13.73 on test set. We have also gathered all model cards for the models that were used for further experimentation.

This study yeilds promising future directions as the experiment was done on only 8 corpora. We suggest to increase the number and observe the derivation of the result. We also suggest to combine Bantu language as one multi-lingual corpora and observe the result. The experiment has been done on a novel Umsuka parallel corpora, the study should extend to more common benchmarks. This study should extend to different low-resourced languages of different continents of our world. We have derived a formula that takes into the account just the distance and the size of corpora, a promising research would be to derive a formula that takes morphologies and/or phonologies and fives a distance based on that. With NGDC at hand, it motivates to create a framework where one enters a target language, a $D_{max}$ and a value for weight coefficient $c$ and gets desirable models to train on. There are many precise ways of finding GD, such as Lambert's formula \citep{lambert1942distance} and Vincenty's formula \citep{vincenty1975direct} which may enhance NGDC's performance. It also opens up ways to introduce morphology in the formula, which we expect it to improve the overall selection of the models.

\bibliography{anthology,custom}
\bibliographystyle{acl_natbib}

\end{document}